\documentclass[10pt,twocolumn,letterpaper]{article}

\usepackage[pagenumbers]{wacv} 

\usepackage{graphicx}
\usepackage{amsmath}
\usepackage{amssymb}
\usepackage{booktabs}

\usepackage{amssymb}
\usepackage{pifont}
\usepackage{xspace}

\usepackage{xr}

\newcommand{\cmark}{\ding{51} \xspace}%
\newcommand{\xmark}{\ding{55} \xspace}%

\usepackage[pagebackref,breaklinks,colorlinks]{hyperref}

\usepackage[capitalize]{cleveref}
\crefname{section}{Sec.}{Secs.}
\Crefname{section}{Section}{Sections}
\Crefname{table}{Table}{Tables}
\crefname{table}{Tab.}{Tabs.}

\def\wacvPaperID{615} 
\def\confName{WACV}
\def\confYear{2025}

\begin{document}

\title{Towards Utilising a Range of Neural Activations for Comprehending Representational Associations 
}

\author{Laura O'Mahony, \hfill Nikola S. Nikolov, \hfill David J.P. O'Sullivan \\
University of Limerick
\\
{\tt\small lauraaisling.ml@gmail.com}
}
\maketitle

\begin{abstract}

    Recent efforts to understand intermediate representations in deep neural networks have commonly attempted to label individual neurons and combinations of neurons that make up linear directions in the latent space by examining extremal neuron activations and the highest direction projections. In this paper, we show that this approach, although yielding a good approximation for many purposes, fails to capture valuable information about the behaviour of a representation. 
    Neural network activations are generally dense, and so a more complex, but realistic scenario is that linear directions encode information at various levels of stimulation. We hypothesise that non-extremal level activations contain complex information worth investigating, such as statistical associations, and thus may be used to locate confounding human interpretable concepts. 
    We explore the value of studying a range of neuron activations by taking the case of mid-level output neuron activations and demonstrate on a synthetic dataset how they can inform us about aspects of representations in the penultimate layer not evident through analysing maximal activations alone. 
    We use our findings to develop a method to curate data from mid-range logit samples for retraining to mitigate spurious correlations, or confounding concepts in the penultimate layer, on real benchmark datasets. 
    The success of our method exemplifies the utility of inspecting non-maximal activations to extract complex relationships learned by models. 
\end{abstract}


\section{Introduction}
\label{sec:introduction}

Many previous interpretability works have gained valuable insights by taking high levels of neuron activations~\cite{olah2017feature, bau2017network, bykov2024labeling} and concept projections~\cite{o2023disentangling, graziani2023uncovering, fel2023craft}. 
Analysing the highest activation levels necessarily removes a level of complexity that makes it manageable to gain insight into representations encoded by neurons, and concepts contained by groups of neurons within layers. 
This coarse-grain analysis has yielded invaluable insights into the nature of neural network representations through finding human interpretable concepts. 
However, models are a function of the data they are trained on, so in reality the representations of concepts are often not disentangled. It is well known that they may not separate the interpretable, independent and informative factors of variations in the data~\cite{comon1994independent, bengio2013representation}. 
Understanding how these factors intersect can inform us of the relationships between concepts in a model and the robustness of its performance. 
The meaning and utility of non-maximal activations of neurons and directions, more generally, have not received the same level of attention as maximal activations. 
Explanations for various levels of neuron activations is a nascent subfield of interpretability research~\cite{la2024towards}, which we will argue is an underused technique to understand model representations. 

This paper focuses on understanding some of the complexities contained in deep network representations that are not easily extracted from maximal activations. 
There are many popular approaches that focus on the highest activation levels, including visualisation and inversion procedures. 
These work by iteratively altering an initially noisy input to obtain an input that maximises a neuron’s activation~\cite{olah2017feature, olah2020zoom, erhan2009visualizing, mahendran2015understanding}. 
Dataset-based approaches are also commonly used. These techniques typically select samples based on which samples activate a neuron or concept vector the most
~\cite{bau2017network, bau2018gan, mu2020compositional, o2023disentangling, graziani2023uncovering}. 
This paper takes the dataset-based approach, where we obtain insights by looking at dataset examples, including samples outside the highest activation range, in order to comprehend representational associations. 
1We study the advantage of this lens in the penultimate embedding layer and the output neurons of a model in a classification setting. 
We show that considering different levels of output neuron activations can not only highlight mislabels and atypical inputs, but can afford insight into the entanglement of concepts allowing us to find and mitigate spurious patterns learned by a model. 

\begin{figure*}[ht]
\centering
    \includegraphics[width=0.85\linewidth]{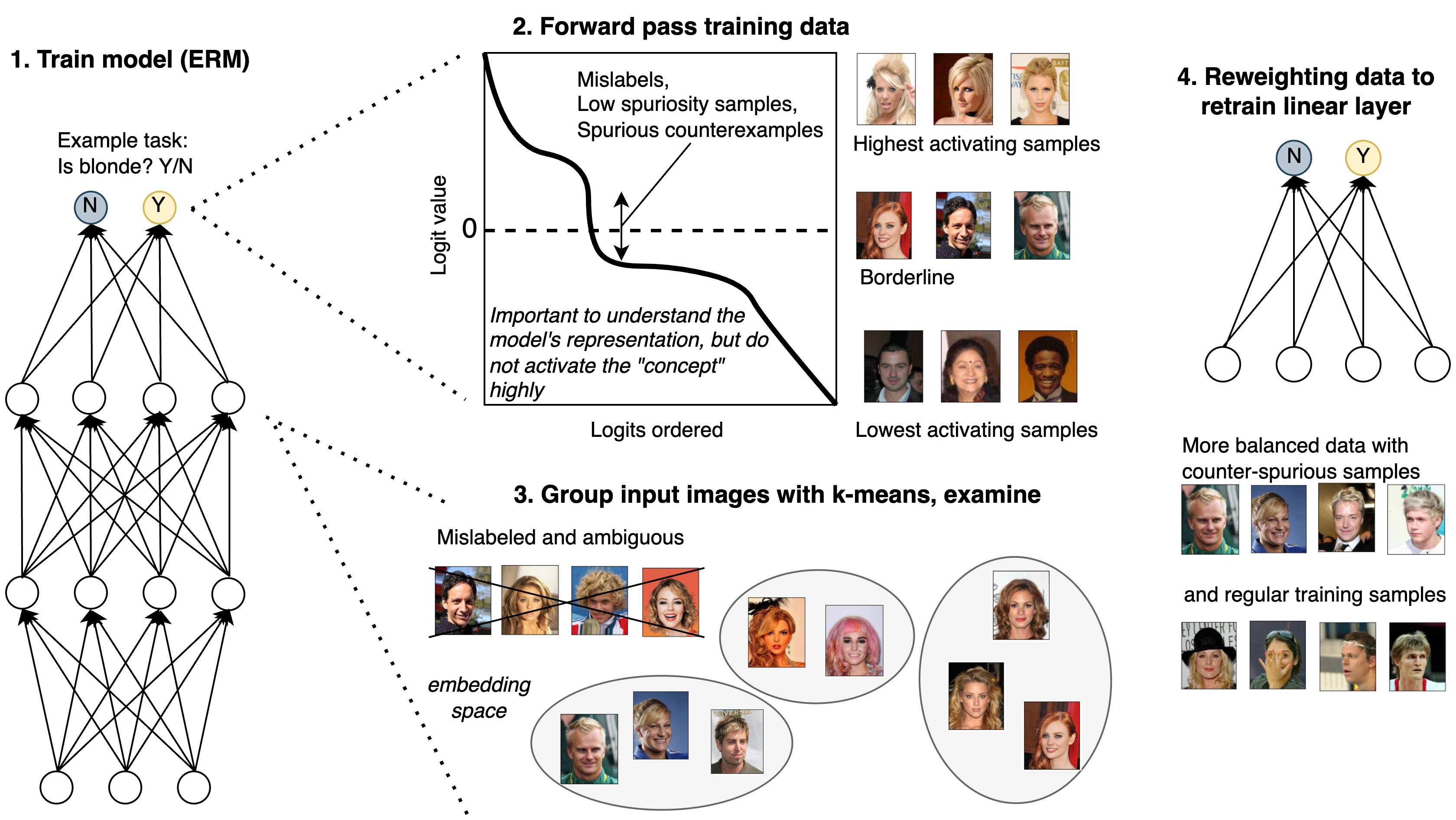}
\caption{An illustration of the spurious correlation data analysis and automatic selection for retraining to mitigate the found spurious bias in the CelebA dataset. 1. A model is trained using the standard empirical risk minimisation (ERM) approach, which typically absorbs spurious correlations. 2. We select an output class and order the logits by magnitude. Samples within the mid-level range where the model has lower prediction confidence in its prediction may include 
many mislabels, low spurious images, and counterexamples to the spurious trend. We filter the data to keep only mid-range activating examples. 3. We cluster the corresponding embeddings from the penultimate layer. 4. In this way, less spurious data is selected to retrain the model, such as by simply fine-tuning the classification layer. 
}
\label{fig:overview}
\end{figure*}

To better understand the extent of the connection between activation levels and confounding concepts, we take the case of output neurons, and first examine harmful spurious patterns - patterns that are correlated with the data class labels but are inherently irrelevant to the task - 
to the data class labels related to shortcuts in classification~\cite{murali2023beyond}. 
This phenomenon of shortcuts permeates the landscape of deep learning~\cite{hermann2020shapes, trivedi2023a}. 
Models learn patterns from their training data and 
may not preferentially use the same semantic features as humans~\cite{geirhos2020shortcut}.
For example, models have been found to take advantage of background cues~\cite{Ribeiro2016} or co-occurring object locations~\cite{geirhos2020shortcut} if such patterns are present in the training data. See further discussion in~\cref{apn:related_work}

We begin with a motivating demonstration of a scenario where inspecting maximal examples fails to highlight a known bias learned by the model, whereas exploring mid-range activating samples reveals samples containing the bias. 
In particular, we consider a toy setting of synthetic data with disjoint sets of core (shape of the object to be classified, i.e., square, circle, oval) and spurious features (location of object, i.e., left, right, or middle of the image). 
Here, in a setting where factors of variation in the data are known to be correlated with the ground truth label, we demonstrate that \textbf{the complexity of deep network representations is not easily extracted from maximal activations}. 
In this setting, we make observations about the nature of spurious correlations 
that inspire our experiments
on two spurious feature benchmark datasets (CelebA~\cite{liu2015deep} and Waterbirds~\cite{welinder2010caltech, xiao2010sun}). 
These experiments serve to present empirical evidence that 
non-extremal activations are useful 
1) to identify mislabeled samples;
2) to characterise the nuances of the representations learned by a model, such as 
a concept's representational relationship to other concepts, and 
3) to select samples useful to finetune a model so that the model becomes less reliant on spurious shortcuts. 
We validate the advantage of exploring mid-range activating examples by identifying counter-spurious data, i.e., groups of data that break the spurious trend, 
and fine-tuning to improve model performance on such minority groups. 
We illustrate our lens and method to identify mislabels, spurious and counter-spurious data, and data we employ to mitigate spurious correlations a model has learned in \cref{fig:overview}, which will be discussed in detail in \cref{sec:mid}.

In this manuscript, we make a step towards utilising the rich information contained in non-maximal levels of neuron activations and concept projections by evidencing the utility of a range of activations in recognising potential confounding concepts and data mislabels. We employ our insights to develop a simple method to identify and mitigate potential spurious correlations. We hope our contributions grant insight into the messy, non-binary nature of neural representations, emphasise the need for examining various levels of the relevant unit of interpretability, and inspire further work in connecting interpretability research and confounding concepts. 
%
Our code will be made publicly available\footnote{\href{https://github.com/lauraaisling/MID-level-activations}{https://github.com/lauraaisling/MID-level-activations}}. 

\section{Related Work} 
\label{sec:related}

\paragraph{Maximal Activations in Mechanistic Interpretability and Concept-based Explanations.} %

Mechanistic interpretability aims to understand models by breaking them down into fundamental units. 
Many of these methods label circuits and neurons in networks via feature visualisation~\cite{olah2017feature,olah2020zoom},
by 
finding samples with maximal concept activations and directions in latent space~\cite{bau2017network,fong2018net2vec, kim2018interpretability, mcgrath2022acquisition, lucieri2020interpretability, zhou2018interpretable, fel2024holistic}, using dictionary learning techniques~\cite{fel2024holistic, templeton2024scaling, gao2024scaling}, or activation patching~\cite{vig2020investigating}. 
To localise an abstract concept to a model component used during inference is not as simple as identifying an injective neuron-concept relationship for each relevant concept, because neurons can be polysemantic – they can be activated by multiple high-level concepts~\cite{smolensky1988proper, rumelhart1986parallel, olah2020zoom, bolukbasi2021interpretability}. 
Concept-based interpretability is the idea of describing models at a higher or conceptual level. 
Describing connectionist models, or artificial neural networks 
using patterns of activity that have conceptual semantics, has been thought to yield a good approximation for many purposes since Smolensky~\cite{smolensky1988proper}. 
Many works have shown the existence of high-level human interpretable concepts such as textures, shapes, and parts of objects present as neurons~\cite{bau2017network,la2024towards} and directions in activation space~\cite{alain2016understanding, fong2018net2vec, kim2018interpretability, zhou2018interpretable, graziani2023uncovering, szegedy2013intriguing, mikolov2013linguistic, bau2017network, elhage2022toy, wang2023concept, park2023linear, trager2023linear}. 

Many interpretability techniques have been proven to be flawed. For example, Geirhos \etal~\cite{geirhos2023don} and Makelov \etal~\cite{makelov2023subspace} showcase the unreliability of feature visualisation and activation patching, respectively. 
Graziani \etal~\cite{graziani2023uncovering} found that maximally activating images are either visually similar or exaggerate the motifs revealed by activation maps. 
La Rosa \etal~\cite{la2024towards} worked on explanations for different levels of neuron activations.
Nicolson \etal~\cite{nicolson2024explaining} found that Concept Activation Vectors (CAVs), or concept vectors calculated with TCAV~\cite{kim2018interpretability} for specified concepts, may be entangled with different concepts, \eg, the concepts blue and sky. The associated CAVs can then lead to misleading explanations. 
It is well known in the disentanglement literature that models do not necessarily capture independent factors of variation in the data~\cite{hsu2024disentanglement, bricken2023towards, chen2018isolating,bengio2013representation,comon1994independent,hyvarinen2000independent} allowing for manual probing of disentangled representations for human-interpretable concepts. 
In this light, our work aims to go beyond locating human interpretable concepts and instead utilise interpretability to provide insights into the dataset and model~\cite{graziani2023concept, nicolson2024explaining}. 
Similar to La Rosa \etal~\cite{la2024towards}, we do not restrict our analysis to maximally activating examples. Our work considers non-maximally activating examples helpful in understanding when models may entangle various factors with class labels. 



\paragraph{Spurious Correlations.}

Models trained via standard empirical risk minimisation (ERM) are optimised for accuracy alone. Hence, they commonly depend on any patterns predictive of class that are present in the training data, even if they are irrelevant to the true label. Models may use spurious features when spurious correlations are easier to learn than core features~\cite{shah2020pitfalls, murali2023beyond}
and therefore use shortcuts~\cite{geirhos2020shortcut}. 
The processes that determine which features a model represents are complex~\cite{hermann2020shapes} (See further discussion in \cref{apn:related_work}). 
A reliance on spurious features is problematic when the data distribution shifts in a way that breaks these correlations, such as in image backgrounds~\cite{xiao2020noise, moayeri2022comprehensive}, 
texture bias~\cite{geirhos2018imagenet, hermann2020origins, islam2021shape}, and OOD data~\cite{hendrycks2022baseline}. 
In addition to performance impacts on non-spurious images, there can be major implications for model fairness~\cite{chouldechova2017fair, buolamwini2018gender, hashimoto2018fairness, khani2021removing, ye2024spurious}. 

Many works seek to improve robustness to spurious correlations by optimising for worst-group accuracy~\cite{hu2018does, sagawa2019distributionally, zhang2020coping, sohoni2020no, liu2021just}. 
A common approach in recent literature is to retrain, focusing on correcting errors~\cite{liu2021just, zhang2022correct}, 
many do this by balancing predominantly spurious data with minority samples~\cite{kirichenko2022last}. 
The requirement for minority group information varies among these methods, with some requiring spurious attribute labels on the entire trainset~\cite{idrissi2022simple}, 
while others propagate minority group information known from the validation data to the training data~\cite{nam2022spread}. 
Few methods require no access to any group labels of the subclasses created due to stratification by the spurious attribute. 
GEORGE~\cite{sohoni2020no} is one such method that estimates group labels using unsupervised clustering and subsequently trains a more robust classiﬁer. 
A recent work develops a human-in-the-loop method of sorting data by spuriosity~\cite{moayeri2024spuriosity}. 
The authors suggest this ranked data could then be used to retrain and mitigate the bias, effectively better utilising the data already available. 
This is because recent work has evidenced that core features are still learned under ERM even when spurious features are favoured. This means simple fine-tuning on data without the spurious feature can efficiently reduce spurious feature reliance, even without full model retraining~\cite{kirichenko2022last, izmailov2022feature}. 
Full retraining methods may or may not suppress spurious features, and such final layer fine-tuning methods act to adjust the final layer feature contributions to the model output prediction~\cite{hermann2020shapes,kirichenko2022last, izmailov2022feature}. 
See further discussion in~\cref{apn:related_work}. 
Our work 
relates to spurious correlation works, as we identify patterns and groups of data where the model exhibits uncertainty without using any group information. Following methods measuring minority group performance, we additionally quantify model worst-group accuracy improvements.

\section{Problem Setup}
\label{sec:setup}


We consider a classification task with a training set $\{(x_i, y_i, s_i)\}^n_{i=1}$ consisting of $n$ training datapoints. Each datapoint $x_i  \in X$ is associated with a class label $y_i  \in Y$, a category the data is classified to belong to.
In addition, the data is stratified by an attribute $s \in S$. In our setting, we do not assume knowledge about $s$, or that we have the labels $s_i$. 
These spurious attributes have no causal relationship to the label and may be an artefact of the data collection processes or an unknown bias in the data. 
Due to such a bias, 
such features may contribute to a model's prediction if they have predictive power towards a certain class. 
Therefore, their relationship to the labels may change between the training and test distributions, meaning a model relying on such a spurious feature will perform poorly at test time. 
Let us define the group associated with each datapoint $x_i$ as a combination of the label $y_i$ and spurious attribute label $s_i$, so each datapoint $x_i$ belongs to some group $g_i \in G$. 
This leads to the majority groups of samples with a positive spurious relationship between the label and spurious attribute largely comprising the training data, while only a small proportion of the total training data consists of minority groups without such a relationship.
For this reason, a model trained via ERM 
may rely on the spurious feature and subsequently have a 
decrease in accuracy for minority groups in the training data instead of basing predictions on core features and achieving uniform accuracies across groups. This gap is commonly measured by calculating the lowest test accuracy across all the groups $G$, known as the worst-group accuracy (WGA). 
Next, we examine a spurious relationship and non-maximal activations on a synthetic dataset where we have the privilege of controlling the amount of bias.

\section{Understanding Representations with Spurious Correlations}
\label{sec:understanding}

\subsection{Synthetic Data}
\label{sec:synthetic_data}

We consider a synthetic dataset that allows us to sample images with disentangled core and spurious features so that the factors of variation can be correlated with the ground truth labels to various degrees. 
The DSprites dataset~\cite{higgins2017beta} consists of 700,000 unique images of size $64 \times 64$ pixels each with a white shape (Square, Oval, Heart) against a black background. 
This dataset was specifically designed to measure how much a learned representation has disentangled the sources of variation - Scale, Orientation, X-Position, and Y-Position (details in \cref{apn:dsprites}) - and has been used extensively in the disentanglement literature~\cite{higgins2017beta,shu2019weakly,locatello2019challenging,locatello2020weakly,van2021neuromatch}.

\subsection{Unfair Classification}
\label{sec:unfair_classification}

In the original dataset, the factors of variation are uncorrelated with the shape. 
However, Creager \etal~\cite{creager2019flexibly} introduced a DSpritesUnfair variant of this dataset for fairness problems, where one is concerned with correlations between attributes and labels. 
We similarly simulate a bias in the training dataset by selecting images with an ``unfair'' sampling procedure where the label (Shape) is correlated with its $x$-position to varying extents. 
Specifically, we set the bias proportion of the samples to have squares on the left, ovals in the middle, and hearts on the right.
See~\cref{apn:dsprites} for further detail. 
The task is to classify a shape as a square, oval, or heart, and the introduction of a bias creates a tendency to make predictions of shapes using their location in the image.

\begin{figure}[h]
\centering
    \begin{subfigure}{\linewidth}
    \centering
    \includegraphics[width=0.6\textwidth]{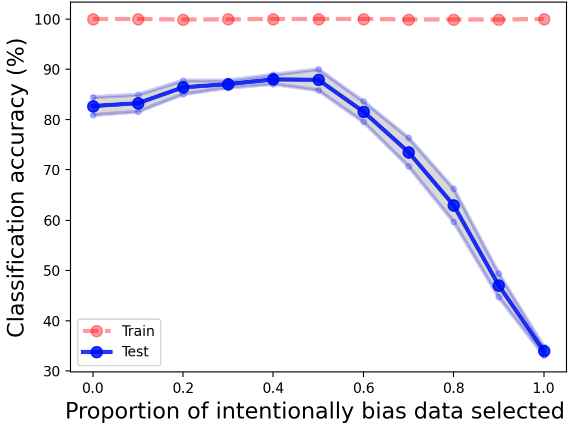} 
    \caption{} 
    \end{subfigure}
\hfill
    \begin{subfigure}{\linewidth}
    \centering
    \includegraphics[width=\textwidth]{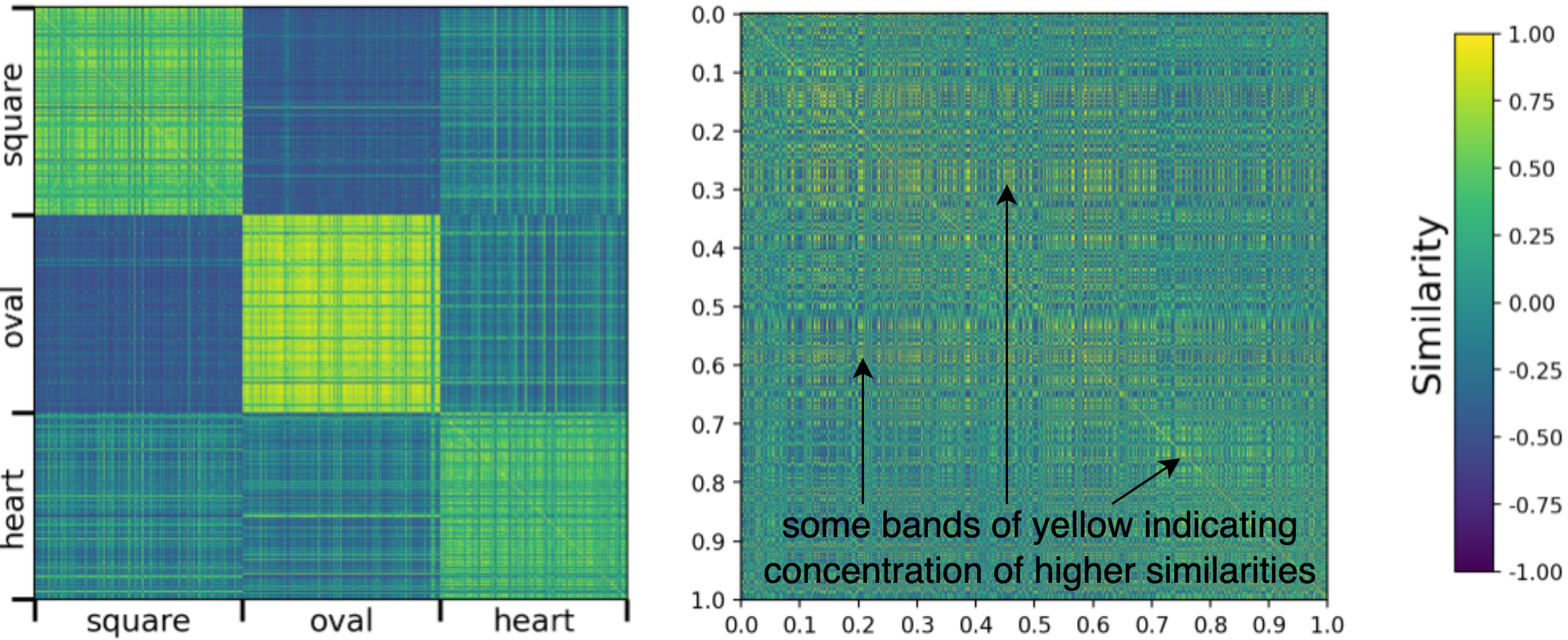} 
    \caption{} 
    \end{subfigure}
\caption{(a) Train and test classification accuracy results on DSprites dataset for classifiers trained with varying proportions of bias added (shown on the $x$-axis). (b) The representation similarity matrix for the penultimate layer for bias level 0.4 sorted by shape (left) and $x$ position (right). 
} 
\label{fig:DSprites_posX_bias_results}%
\end{figure}

To do this task, we use a 
convolutional neural network consisting of two convolutional layers followed by three fully connected layers and a final classifier layer predicting the shape. 
We find this architecture trained on an unbiased dataset can easily classify novel test samples. 
We train models to perform well on training datasets with varying proportions of the simulated bias and calculate test accuracies on a fair test dataset, meaning there is no relationship between the position of the shape and the label. 
\cref{fig:DSprites_posX_bias_results} (a) shows mean accuracies (over five seeds) and 95\% confidence intervals for models trained on each dataset of varying proportion of bias for the training and test sets. 
Unsurprisingly, the model 
performance on the training data is high for each level of bias, but the test performance reduces and approaches random when the model relies on the position shortcut learned from the training dataset for high levels of bias where the positional shortcut has predictive power. 
We see that, for models trained with bias levels 0.1-0.5, 
the weak spurious correlation helped the model perform better on a fair test set. We conjecture that this could be related to the increased signal provided by the spurious signal interacting with the loss landscape. The spurious correlation hurts performance on a test set when this spurious signal overpowers the core features.
See~\cref{apn:murali} for a discussion on ``harmful" spurious correlation, and this seemingly ``helpful" one.

For many tasks and use cases, we may not want models to rely on co-occurring attributes, and for numerous applications, it is essential to at least understand if a model is influenced by co-occurring attributes to the label.
For illustration purposes, we focus on bias level 0.4 (corresponding to the position shortcut for classification holding for $\sim$ 60\% of the training data) and conduct representational similarity analysis using cosine similarity as a measure of representational similarity. 
We include further demonstrations for other bias levels in~\cref{apn:dsprites}.
\cref{fig:DSprites_posX_bias_results} (b) shows representation similarities of the encoder (last fully connected layer) representations between each pair of datapoints for 1000 test examples. 
This matrix of similarities, sorted by the shape (label), is given on the left. 
For good representations, we should see a high representation similarity for the diagonal blocks (examples of the same class) and low similarity for the other blocks. 
We see that the model represents ovals and squares very similarly to other images of the respective class, and distinct from the other class. The model's representations for the heart class generally exhibit a higher intraclass than interclass similarity but are somewhat muddled with the other two classes. 
The similarities sorted by $x$ position (the bias attribute) are illustrated on the right of the subplot. 
The patterns of similarity are substantially more random than for shape, but 
small bands of yellow can be seen as annotated. This is in contrast to the absence of any pattern for the 0.0 bias level shown in Fig.~\ref{fig:RSMs} in the appendix.  
This means that the model, to some extent, encodes the position attribute, not just the shape. 
We would expect this for levels of bias that negatively impact model accuracy, but this example shows that the model uses position in its representation 
not exclusively for ``harmful'' bias levels. 
Refer to \cref{apn:dsprites} for further demonstrations of \cref{fig:DSprites_posX_bias_results} (b) for various bias levels. 

\begin{figure}[ht]
\centering
    \begin{subfigure}{.45\linewidth}
    \includegraphics[width=\textwidth]{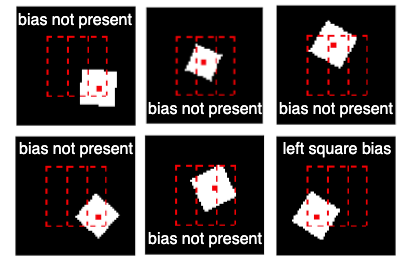} 
    \caption{} 
    \end{subfigure}
\hfill
    \begin{subfigure}{.45\linewidth}
    \includegraphics[width=\textwidth]{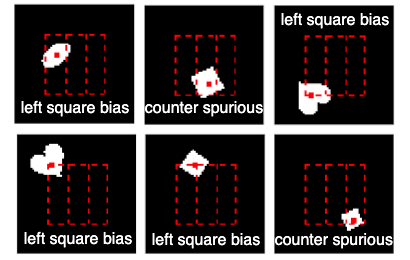} 
    \caption{} 
    \end{subfigure}
\caption{
(a) Maximally activating examples for the neuron corresponding to the square class for training bias level 0.4. 
(b) Mid-level activating images in the same setting.} 
\label{fig:DSprites_logit_images}
\end{figure}

In \cref{fig:DSprites_logit_images}, we demonstrate some observations for this same bias level. 
The logit values are the raw class scores output by the final layer of the neural network before applying an activation function.
We rank the images in descending order using their logit value for the class square. 
In \cref{fig:DSprites_logit_images} (a), we plot the top 6 scoring images, or maximally activating examples. 
Note how the highest activating squares 
seem to all have a similar large size, but notably do not reflect the position spurious attribute. 
We also 
show a random sample of 6 images from around the point where the logit for the square class changes in sign in \cref{fig:DSprites_logit_images} (b).
Plotting these samples shows 3/6 shapes that are not square but are located in the biased region (left of the image), 1 square on the left, and we have counter-spurious squares as they are squares but are not located in the left region the model associates with square. 

We note that this pattern from \cref{fig:DSprites_logit_images} 
is shown for lower bias levels where the bias did not negatively impact the model test accuracy (it was, in fact, positively impacted). 
As expected, this also holds for harmful bias levels as shown in the appendix. 

\section{Retraining to Mitigate Spuriosity}
\label{sec:spurious_retraining}

In \cref{sec:understanding}, we demonstrated that even models trained via standard ERM that achieve 100\% training accuracy contain additional information that can be gleaned from the classification logits. 
We have seen that studying samples that do not activate a ``concept'' highly are important to understand a model's representation. 
Inspired by these observations, we propose a practical method \textit{Modifying with logit Intercept Data} (MID) for reducing a model's reliance on spurious correlations.

\subsection{MID - Modifying with Intercept Data}
\label{sec:mid}

MID is a simple approach that involves modifying an ERM model using a fine-tuning approach with data on which the model doesn't perform well, breaking a spurious trend. The approach comprises four steps, summarised in \cref{fig:overview}.
Step 1 is training a model $M$ via standard ERM on a dataset $D = \{x_i,y_i\}$. We assume the distribution of $x_i$ corresponding to $y_i$ emits some spurious correlation, meaning some grouping may be contributing as a shortcut to the classification task. 
We consider the network up to the penultimate layer as an encoder or feature extractor (\eg, convolutional or transformer layers) with a multi-layer perceptron classification layer that maps to logits. In step 2, we select a class $y \in Y$ and order the logits of this class. We choose the images with a low magnitude logit, images near the 0 logit, or $x$ ``intercept" in a plot of the logits ordered as in \cref{fig:overview}. The highest logits are the maximally activating examples to the class. Whereas these middle logits are samples the model is less confident about for some yet unidentified reason. 
Many of these images could be low-quality samples, mislabels, as well as low spuriosity data, that is, data belonging to the class but not containing the spurious attribute associated with the class, and ``spurious counterexamples", meaning they contain spurious cues that are associated with the class but are not part of that class. We keep only examples from the mid-range of activations for each class and concatenate these. Step 3 involves taking the feature maps extracted by the encoder for this ``intercept data". We first check for mislabelled or ambiguously labelled data, a common occurrence in many datasets~\cite{pleiss2020identifying, long2024understanding}. This could be done using a multimodal model for captioning or Visual Question Answering (VQA) model such as BLIP~\cite{li2022blip}, or using a domain expert for domain-specific data. In our pipeline, we use BLIP in a VQA setting. Further details are given in \cref{apn:filtering}. Following this, we remove data with disputing labels and cluster the remainder of the intercept data using k-means. 
We start with $k=2$ and increase $k$ until a pattern emerges. As we do not assume knowledge of, or access to the spurious attribute labels $s_i$, a brief manual inspection of the images is undertaken to identify reasons why a cluster may not perform well, \eg, in CelebA, we find valuable results with $k=3$. Results show that one cluster had a hair colour that was between brown and blonde, and another cluster had unusual images with artefacts such as poor lighting. In both of these clusters, it is reasonable to expect a model to have a lower prediction confidence. However, a third cluster consisted of blonde people with masculine-associated features, for example, jaw shape and short hairstyles; this shows a harmful spurious pattern that the model has learned, where feminine features are associated with the class blonde and vice versa. 
Full details on the k-means implementation and clustering results are also given in \cref{apn:clustering}. 
The final step involves fine-tuning the model with balanced data obtained from the clustering step. Retraining the model from scratch by upweighting the examples countering the spurious trend is possible. However, Kirichenko \etal~\cite{kirichenko2022last} find evidence that neural networks learn high-quality representations of the core features relevant to the problem even if they rely on spurious features. Therefore, we can simply and cheaply retrain the classification layer to drop spurious features using data countering the spurious trend using logistic regression as per DFR~\cite{kirichenko2022last}. 

\section{Experiments }
\label{sec:experiments}

In this section, we discuss the implementation of MID, 
our method for identifying groups of data the model is uncertain about and retraining the model to perform better on spurious groups. We evaluate MID on benchmark problems with spurious correlations and comparable methods.

\subsection{Data}
\label{sec:data}

We study datasets observed to have poor worst-group accuracy (WGA) due to spurious correlations. We briefly describe the task and spurious attribute in the training set and reserve further details for \cref{apn:datasets}. 

\noindent
\textbf{Waterbirds}
is a common benchmark binary classification task of distinguishing between land birds and water birds. The dataset contains a background spurious correlation as it combines foregrounds from Caltech Birds~\cite{welinder2010caltech} and backgrounds from SUN Places~\cite{xiao2010sun} with a majority 
consisting of land birds over land backgrounds (73\%) and water birds over water backgrounds (22\%), and a minority breaking this spurious correlation (4\% and 1\%). The test set breaks this shortcut. 

\noindent
\textbf{CelebA} 
hair colour prediction~\cite{liu2015deep} tasks us to classify between images containing blonde and non-blonde celebrities. The training data contains predominantly non-blonde females (44\%), non-blonde males (41\%), and blonde females (14\%) with a minority group of blonde males (1\%), making gender a spurious feature. Again, this shortcut does not generalise well to the test set.

\subsection{Baselines}
\label{sec:baselines}

We compare MID with the base empirical risk minimisation (ERM) model from Step 1, as well as many state-of-the-art baseline methods. We compare to Just Train Twice (JTT)~\cite{liu2021just}, a method that detects the minority group examples by identifying examples with high loss during initial training. 
Group labels in the validation data are used to tune hyper-parameters for complete retraining with upweighting of the selected examples. 
Correct-n-Contrast (CnC)~\cite{zhang2022correct} similarly detects the minority group examples and learns representations robust to spurious correlations using a contrastive objective. 
Deep Feature Reweighting (DFR)~\cite{kirichenko2022last} retrains the last layer of the model trained via ERM using a group-balanced subset of the data, which assumes group labels. We take this as an upper bound of how well our method could do if we had labelled training group data. 
Spread Spurious Attribute (SSA)~\cite{nam2022spread} uses a semi-supervised approach that propagates the group labels known from the validation data to the training data.
SUBG~\cite{idrissi2022simple} is equivalent to ERM applied to a random subset of the data where the minority groups are equally well represented. 
The closest method to ours is GEORGE~\cite{sohoni2020no}, the only other method besides ERM requiring no prior information about the groups. It estimates the group labels using unsupervised clustering and uses these estimates to train a robust classiﬁer. 

\subsection{Hyper-parameters}
\label{sec:hyper-parameters}

We use a ResNet-50~\cite{he2016deep} pretrained on ImageNet~\cite{deng2009imagenet, russakovsky2015imagenet} for Waterbirds and CelebA classification tasks, as this architecture is used in previous works (\eg, ~\cite{liu2021just, kirichenko2022last, izmailov2022feature, nam2022spread, sohoni2020no}). 
We finetune the ImageNet pretrained models for 100 epochs and 50 epochs respectively on one NVIDIA A100 GPU. Further details on ERM training are documented in \cref{apn:datasets}. 
For retraining, we fine-tune only the classification layer, as is found by Izmailov \etal~\cite{izmailov2022feature} that ERM models learn core features on datasets with spurious correlations despite the model relying on spurious features to make predictions. Similar to Kirichenko \etal~\cite{kirichenko2022last}, we apply $\ell_1$ regularisation to allow the model to drop the spurious features. We tune the strength of this regularisation parameter without assuming access to a validation dataset with group labels. We note that most of the prior work methods including JTT, CnC, SSA, and SUBG tune hyper-parameters on the validation data, we take DFR$^{Tr}_{Tr}$ as a method that tunes hyper-parameters on the training data (with group data known). 
In MID, we split the retraining data in half and use this to train the regularisation hyper-parameter. We then train the classifier using logistic regression with this optimised hyper-parameter on the full retraining subset of training data. 
We train this final linear layer atop the frozen feature extractor for each of our models on the CelebA and Waterbirds datasets for ten epochs. 
We implement this using logistic regression 10 times using different random subsets of the data, and average the weights and coefficients of the learned models, to regularise and make the best use of the retraining data as was also done by Kirichenko \etal~\cite{kirichenko2022last}. See \cref{apn:clustering} for specific details on retraining. 

\subsection{Results}
\label{sec:results}

The results for MID and the baseline methods described in \cref{sec:baselines} are detailed in \cref{tab:benchmarks}. 
We document the requirement for group information for each method for the train and validation sets. 
We do not assume that we have access to the group labels concerning the spurious attributes for training or validation data. This information is used only for evaluating the WGA. 
We additionally do not require group-balanced validation data, as is assumed by many methods as shown in the table. 
MID is competitive with these methods requiring labelled validation or training group data. MID far exceeds the WGA of the other methods that do not use group labels. It improves the WGA of the base model from step 1 of the method by a large amount across both datasets. It also sees an increase in WGA from GEORGE, the method closest in idea and assumptions to MID. 
Overall, MID performs very well in mean accuracy across groups as well as WGA, without requiring group labels for the training or validation data.

\begin{table}[th]
\begin{center}
\begin{small}
\begin{tabular}{lccccccr}
\toprule
\textbf{Method} & \multicolumn{2}{c}{\textbf{Group Info}} & \multicolumn{2}{c}{\textbf{Waterbirds}} & \multicolumn{2}{c}{\textbf{CelebA}} \\
\cmidrule(l){2-3}
\cmidrule(l){4-5}
\cmidrule(l){6-7}
 & Train & Val & WGA & Mean & WGA & Mean \\
\midrule
JTT & \xmark & \cmark &86.7 &93.3 &81.1 &88.0 \\
CnC & \xmark & \cmark &88.5 &90.9 &88.8 &89.9 \\
SSA & \xmark & \cmark \cmark &89.0 &92.2 &\textbf{89.8} &92.8 \\
SUBG & \cmark & \cmark &89.1 &- &85.6 &- \\
DFR$^{Tr}_{Tr}$ & \xmark \cmark & \xmark &\textbf{90.2} &97.0 &80.7 &85.4 \\ 
\midrule 
GEORGE & \xmark & \xmark &76.2 &95.7 &53.7 &94.6 \\
ERM & \xmark & \xmark &68.4 &\textbf{\underline{98.1}} &47.8 &\textbf{\underline{95.4}} \\
MID & \xmark & \xmark &\underline{87.8} &97.3 &\underline{85.5} &90.9 \\

\bottomrule
\end{tabular}
\end{small}
\end{center}
\caption{
Worst-group accuracy and mean test accuracy of baseline methods, ERM base model, and MID on benchmark datasets. We follow prior works for mean accuracy by weighting the test group accuracies according to their prevalence in the training data. For each method, we identify the requirement for group labels for both the training and validation datasets. Most methods require either training or validation group labels; some require labelled validation data to do hyper-parameter tuning as well as train the model (indicated with \cmark \cmark). MID is competitive with state-of-the-art without requiring labelled group information and is neither weak in terms of WGA nor accuracy. We only use the group information during evaluation. We underline the best-performing method out of those requiring no group labels. 
}
\label{tab:benchmarks}
\end{table}

\section{Discussion}
\label{sec:discussion}

Our experiments show that studying samples that do not activate the ``concept'' highly is crucial to understanding a model's representation, and that examining non-maximal activations has practical utility. 
With the MID framework, the points selected right above the zero logit are more likely to be low spuriosity points memorised by the model, while the images right below the zero logit are more likely to be spurious counterexamples, not belonging to the selected class, but 
containing spurious cues that are associated with the class. 
The selected data may be scanned by a domain expert to understand the reasons for the model's uncertainty on specific data points or clusters of data, or this procedure could be automated for some applications. 

\begin{figure}[h]
\centering
    \begin{subfigure}{.49\linewidth}
    \includegraphics[width=\textwidth]{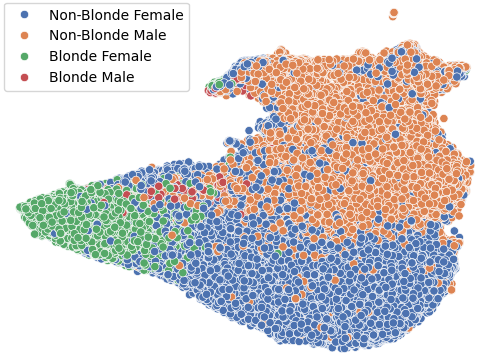} 
    \caption{} 
    \end{subfigure}
\hfill
    \begin{subfigure}{.49\linewidth}
    \includegraphics[width=\textwidth]{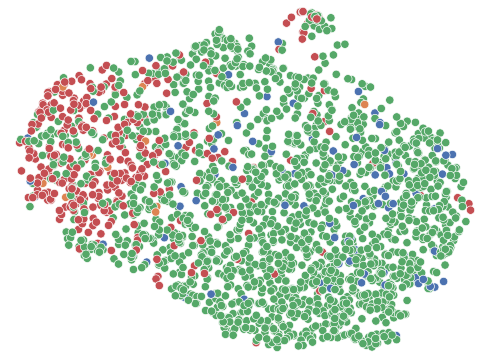} 
    \caption{} 
    \end{subfigure}
\caption{UMAP projections for encoder embeddings of (a) all points and (b) middle logit points from the CelebA dataset. 
} 
\label{fig:UMAPs}%
\end{figure}

In \cref{fig:UMAPs}, we depict the encoder embeddings of datapoints from CelebA, extracted from the ERM model from step 1. (a) shows a UMAP projection~\cite{mcinnes2018umap} of all points in the CelebA dataset. As can be expected from the summary statistics from the CelebA dataset, the minority groups (plotted in red) are largely obscured by the majority ones, and meaningful k-means clusters reflecting spurious patterns could not be found (see~\cref{apn:ablations}). 
(b) displays the UMAP projection for the mid-logit dataset points. Most of these points are female celebrities with a hair colour between blonde and brown, images with poor lighting and low-spuriosity images. We observe a much more apparent signal for points belonging to the minority group allowing meaningful clusters to form. 
This reveals where our method improves on the results from the clustering in GEORGE. GEORGE differs from MID as it attempts to capture signal about minority group datapoints by clustering the UMAP reduction of the unfiltered embeddings using k-means. MID filters the data in a way that narrows down to samples where the model exhibits low prediction confidence, as it may have memorised low spuriosity datapoints and spurious counterexamples to reduce the loss, as in step 2. 
This naturally reduces the amount of data from majority groups that the model has learned well and gives us more minority group data that the model is more uncertain about. 
The appendix documents further details of the method, including 
an analysis of the composition of the clusters in \cref{apn:clustering}.

Our method centres around utilising a model's representations to identify patterns of potential spurious class associations. 
We take output classification neurons as the simplest example of human-interpretable concepts. This has benefits, such as trivial calculation of concept vectors (output neurons) and concept labels already given (in the form of data labels). The concept is, therefore, indisputable. 
Additionally, applying this analysis to study spurious correlations to class labels allowed us to quantify model WGA improvements through probing only mid-level activations. 
However, our analysis of classes is part of a superset of scenarios where we can utilise mid-level activations to locate confounding or entangled concepts within models. 
With minor alteration, our methodology can be applied to study concepts in the latent space of a model found through the state-of-the-art methods for locating concepts in latent layers such as TCAV~\cite{kim2018interpretability}, SVD-based Latent
Space Decomposition~\cite{graziani2023uncovering}, and CRAFT~\cite{fel2023craft}.

We believe that this type of inquiry is underrepresented and underutilised compared to approaches that use maximally activating examples to interpret neurons and concept vectors, as well as methods focused on locating human interpretable concepts. 
It blends well with recent findings that Concept Activation Vectors (CAVs), calculated with TCAV~\cite{kim2018interpretability} for specified concepts, may be entangled with different concepts~\cite{nicolson2024explaining}. Nicolson \etal~\cite{nicolson2024explaining} subsequently recommend verifying expected dependencies between related concepts, a sensible recommendation for confounding concepts that may be located with a method similar to the former steps in MID. 
We leave the study of mid-range activations (or projections in the case of concept vectors) in latent layers of a model to future work. 
We are particularly excited by further works that apply similar investigations to locate confounding concepts in latent layers that may encode concept biases. Extending our work to generative models is a promising application, as they are prone to absorbing bias in training sets. Investigating whether a relationship exists between lack of concept disentanglement and leakage in concept bottleneck models is further compelling~\cite{koh2020concept, mahinpei2021promises, margeloiu2021concept}. 

A potential limitation of our framework is the involvement of a human who is given concise insights into model patterns by viewing some images from the clusters of data the model is uncertain about, and the agency this human has to decide whether or not to remove a located spurious cue. Future work could investigate automating the cluster interpretation step. However, the amount of human involvement is tractable as it is limited to viewing a few images from each cluster. We believe that the human-in-the-loop aspect affords the practitioner greater insight into the model and that verifying expected dependencies between related concepts is crucial in order to trust a model. 

\section{Conclusion}
\label{sec:conclusion}

In this paper, we show a practical use of examining non-maximal activations in locating entangled concepts and understanding the nuances of representations. 
Specifically, we take the case of mid level logit activations and identify patterns of spurious cues, 
instances of inappropriate entanglement of concepts contributing to the classification, 
not made visible by inspecting maximally activating examples. 
In a pilot experiment using synthetic data, we saw that by looking at the representations of examples that do not maximally activate the classification logits, we can better find examples belonging to minority subgroups of the data that the model has not learned well. 
This evidence was solidified by the success of a simple method MID, 
where we locate spurious correlations
on real datasets.
We quantified improvements in model performance on minority groups that can be gained by retraining on our data curation procedure. 
MID is a simple and cheap data identification and model retraining method. It considerably improves the worst-group accuracy of the models trained on benchmark spurious correlation datasets and substantially beats the current state-of-the-art method that requires no labelled group information denoting groups of spurious and non-spurious data.

\section*{Acknowledgements}
\noindent 
LOM acknowledges the support of Science Foundation Ireland under Grant Number 18/CRT/6049. 
We would like to express gratitude to the anonymous reviewers for their helpful comments. 




{\small
\bibliographystyle{ieee_fullname}
\bibliography{egbib}
}

\newpage
\clearpage

\appendix
\section{Appendix}

\subsection{Further Discussion on Related Work}
\label{apn:related_work}

Machine learning does not prescribe what features a model learns, further models may not learn the same features as a human, e.g., models may use texture over shape to classify objects unlike humans~\cite{geirhos2018imagenet}. \textbf{Spurious correlations highlight the importance of understanding what features are learned and how they are represented}. Hermann et Lampinen~\cite{hermann2020shapes} study the relevant question of what shapes feature representations, including an analysis of correlated features, with experiments on a synthetic ``trifeature” dataset of a similar type to our pilot experimental data. 
By measuring the linear decodability of visual features from intermediate model layers, they find that when multiple features redundantly predict class labels, models preferentially represent the feature that is most linearly decodable from the untrained model, hypothesising that the ``decodability of features from an untrained model reflects the model’s inductive biases, and might predict the extent to which a feature would be preserved after training the model on a different task.” There are some theoretical works also backing up their empirical findings~\cite{shwartz2017opening,saxe2013exact,saxe2019mathematical}. 

The authors further find that across training, task-relevant features are enhanced, and irrelevant features for the task are partially suppressed. 
Their findings suggest that the features models represent depend on both the predictivity of features and their easiness of learning. This may explain why more reliable features can be suppressed by a less reliable, but easier-to-learn one. 
With regards to harmful spurious correlations~\cite{murali2023beyond}, this would indicate that a model learns to rely on a less predictive feature since its easiness-to-learn dominates over its lack of reliability for prediction. 
Methods like MID, or other methods that finetune a model relate to these findings as they aim to make the model less sensitive to spurious features. 
They make the easier-to-learn, spurious attribute less predictive by training the model on examples that break the spurious trend. 
They may not (in particular, retraining only the final layer \textit{will} not) reduce the linear decodability of the spurious attribute, but will adjust the feature contributions to the model output prediction.

Kirichenko \etal~\cite{kirichenko2022last} advocate for simple last layer retraining, demonstrating that it can match or outperform state-of-the-art approaches on spurious correlation benchmarks, with far less expense. 
They further show that they could reduce reliance on background and texture information on large ImageNet-trained models using this technique. 
Izmailov \etal~\cite{izmailov2022feature} study feature learning in the setting of spurious correlations. 
They show features learned by simple ERM are highly competitive, and that retraining the last layer beats specialised group robustness methods for reducing the effect of spurious correlations. 
Following this work, in our experiments, we retrained only the last layer. However, other methods of fine-tuning are also possible. 
Final layer retraining alters the weights of how features contribute to the output classification in enforcing the utilisation of semantically consistent features. As mentioned above, this will not alter the decodability of relevant or irrelevant features. However, further layer retraining may or may not suppress the spurious and now less predictive features~\cite{hermann2020shapes} or induce a kind of catastrophic forgetting~\cite{mccloskey1989catastrophic, zenke2017continual}. 

The work of Hendrycks and Gimpel~\cite{hendrycks2022baseline} is also relevant to discuss in relation to our work. The authors observe that accurately classified examples usually have a greater maximum softmax probability than erroneously classified and out-of-distribution (OOD) examples. They use this finding to develop a simple baseline that utilises probabilities from the model’s softmax distribution for detecting if an example is misclassified or out-of-distribution. 
Our work has a different underlying motivation to this work, as we examine logits (model outputs before any activation function is applied), and frame our work as a flavour of interpretability since we are motivated to comprehend representations with a richer level of detail than previous works examining maximally activating examples.
However, our experiments on utilising the mid-range activations on spurious benchmark problems have a similar focus to the goal of Hendrycks and Gimpel as we locate low spurious images (an aspect OOD), and counterexamples to the spurious trend. 
A further overlap with this work is where we found 
that mid-level activations were useful for finding misclassified examples, as these are examples where the model has a near zero output logit value due to exhibiting uncertainty on the example or memorisation of the label. 
An interesting future research direction could be to measure how well mid-level activations can locate out-of-distribution data in these same benchmarks used by Hendrycks and Gimpel.

\subsection{Synthetic Data Training and Experiments }
\label{apn:dsprites}

The available combinations of the DSprites latent dimensions to sample values are described in \cref{tab:DSprites_description}. 

\begin{table}
  \centering
  \begin{tabular}{lll}
    \toprule
    Latent L & n(L) & Values \\ 
    \midrule
    Shapes & 3 & S, O, H \\
    Scales & 6 & 0.5 - 1 \\
    Orientations & 40 & 0 - 2  \\
    Positions in X & 32 & 0 - 1 \\
    Positions in Y & 32 & 0 - 1 \\
    \bottomrule
  \end{tabular}
  \caption{DSprites space of latent dimension values. 
  }
  \label{tab:DSprites_description}
\end{table}

\paragraph{DSpritesUnfair Details:}
The DSprites data consists of 3 shapes (the attribute to be classified), which are homogeneously dispersed in data that is randomly sampled, including by $x$ position. 
Therefore, we would expect each shape to be located in the left, middle, and right segments of the images one-third of the time. 
For each level of bias, $b$, we replace that proportion of the data with data containing only squares to the left, ovals in the middle and hearts on the right, the remaining 1-$b$ proportion of the data is randomly sampled. For example, if 10\% of the data is intentionally biased, the remaining 90\% of the data is the original fair data, so the shortcut holds in 40\% instead of 33.33\% of the data. 

\paragraph{Training Details:}
All models were trained with the Pytorch library~\cite{paszke2019pytorch} with a learning rate of 1e-3, Adam optimiser and a batch size 1000. The loss is the typical cross-entropy loss. DSprites experiments were performed locally. 

\begin{figure*}[h]
\centering
    \begin{subfigure}{\linewidth}
    \centering  
    \includegraphics[width=\textwidth]{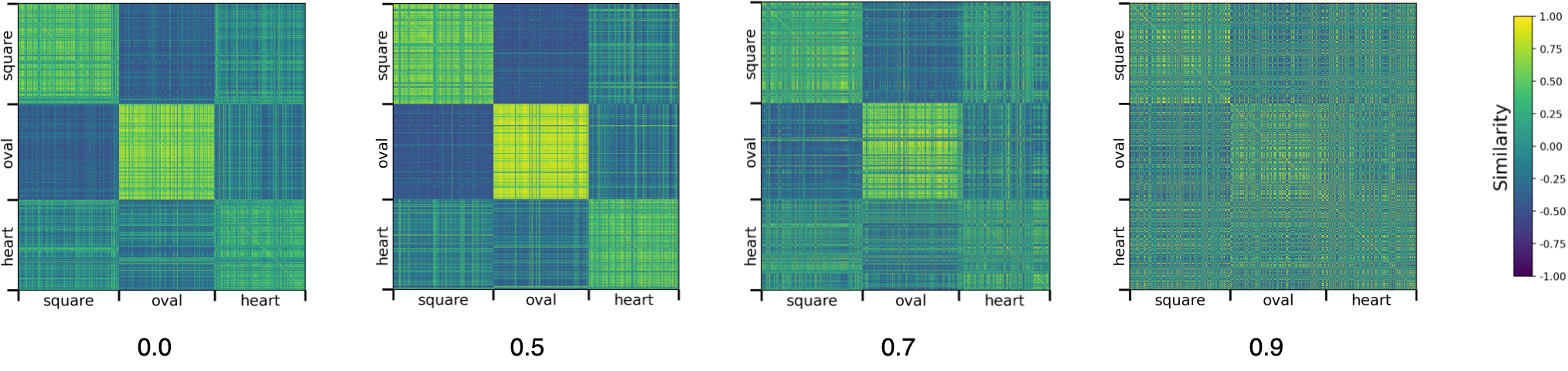} 
    \caption{} 
    \end{subfigure}
\hfill
    \begin{subfigure}{\linewidth}
    \centering
    \includegraphics[width=\textwidth]{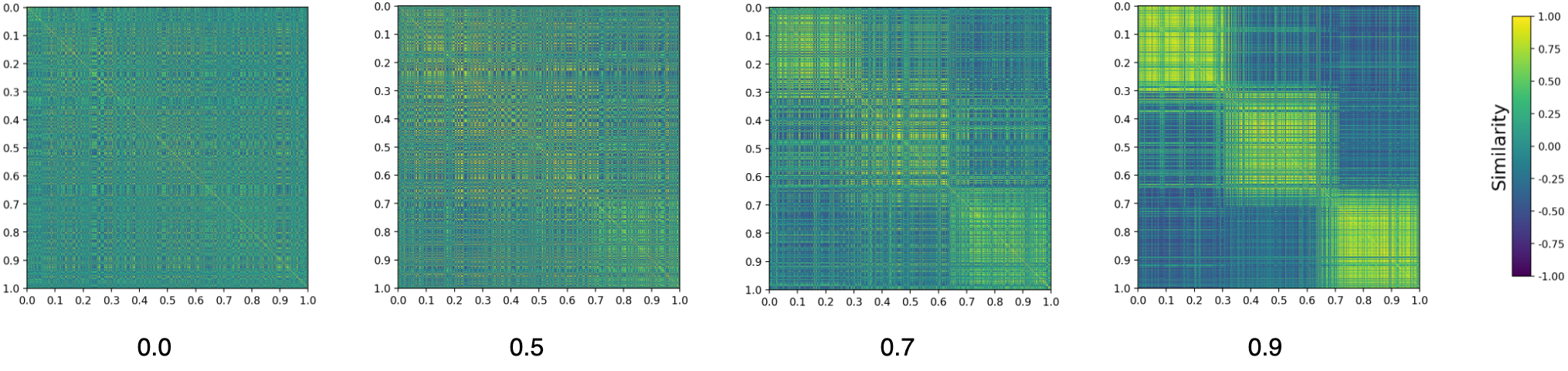} 
    \caption{} 
    \end{subfigure}
\caption{Representational similarity matrices plotted for encoders trained with varying levels of bias (below each image). 
The order of encoder embeddings is sorted by (a) shape, and (b) the $x$ position of the shape. 
} 
\label{fig:RSMs}%
\end{figure*}

\begin{figure}[ht]
\centering
    \begin{subfigure}{.45\linewidth}
    \includegraphics[width=\textwidth]{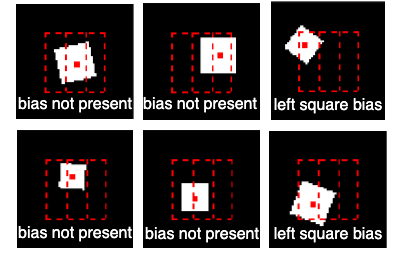} 
    \caption{} 
    \end{subfigure}
\hfill
    \begin{subfigure}{.45\linewidth}
    \includegraphics[width=\textwidth]{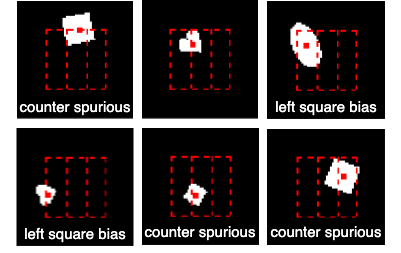} 
    \caption{} 
    \end{subfigure}
\caption{
(a) Maximally activating examples for the neuron corresponding to the square class for training bias level 0.9. 
(b) Mid-level activating images in the same setting.}
\label{fig:DSprites_logit_images_0.9}
\end{figure}

\paragraph{Further Demonstrations of Representational Similarities:}
In \cref{fig:RSMs}, we plot the representation similarities for encoders trained on varying levels of biased training data. Cosine similarity is used as the measure of similarity. For a test sample of 1000 images, the similarity of the encoder's representation for each pair of images is calculated. The images are then sorted by (a) shape, and (b) $x$ position to reveal patterns related to those attributes in the representational space. 
We see that for an encoder trained without bias (level 0.0), the encoder representations for elements of each shape are more similar to other images containing that same shape (high similarity along the diagonals) and the similarities ordered by position show no pattern. For bias level 0.5 (the best-performing model on the test set), the similarity pattern sorted by shape is even clearer. However, some pattern is also present for pairs ordered by position. For bias level 0.7 (where the model's performance on a fair test set is compromised), the pattern in the similarity of representations sorted by shape is not as clean. The pattern in the similarities sorted by $x$ position shows more structure. 
By bias level 0.9 (here, the model's test accuracy is near random as the model relies mostly on the position shortcut), the similarities sorted by shape lack the clear pattern shown for lower bias levels, and a clear pattern can be seen when ordering images by $x$ position. 
\cref{fig:DSprites_logit_images_0.9} shows sample images in the maximal and middle logit range for the class shape ``square" for a model trained on data containing a harmful bias level of 0.9.

\subsection{Harmful and Seemingly ``Helpful" Spurious Correlations}
\label{apn:murali}

Murali \etal~\cite{murali2023beyond} found that when the spurious feature is easier to learn than the core (which they term a \textit{harmful} spurious feature), the model learns to leverage them and that in most cases, the core-only test accuracy drops to nearly random chance. However, results on one dataset (KMNIST with a patch shortcut) showed when a harmful spurious feature was removed in testing, the accuracy dropped, although it remained considerably higher than random accuracy, and the model performance had a wider variance across random seeds. 
In our setting, we found that some relatively small amounts of bias in the train set positively impacted model performance on a fair test set. 
We conjecture that this could be caused by the spurious signal interacting with the loss landscape in a way that makes it easier to find a lower loss. The easier-to-learn spurious pattern provides further signal for the model to find a better local optimum with gradient descent. 
The bias becomes a ``harmful'' spurious correlation (as defined by Murali \etal~\cite{murali2023beyond}) when the spurious signal overpowers the core features.

\subsection{Further details on Spurious Correlation Datasets and Training}
\label{apn:datasets}

\cref{tab:datasets} gives specific details on the compositions of the spurious benchmark datasets used for our experiments. As captured by the ``Group Counts" column, the datasets contain large group imbalances. Further, by conditioning on spurious attribute value, we see that the group imbalances are highly correlated with the labels. In the Waterbirds dataset, it is unlikely to find land birds on a water background. In the CelebA dataset, there are not many images of blonde celebrities who are male. 

Following other works, for CelebA we finetuned the ImageNet pretrained model with stochastic gradient descent using an initial learning rate of 1e-3, a cosine learning rate scheduler, and a weight decay of 1e-4 for 50 epochs. 
For Waterbirds, we set an initial learning rate to 3e-3 and trained for 100 epochs, all else the same.

\begin{table*}[th]
\begin{center}
\begin{small}
\begin{tabular}{lcccccccr}
\toprule
\textbf{Dataset} & \textbf{Labels} & \multicolumn{2}{c}{\textbf{Group Counts}} & \textbf{Class Totals} & \multicolumn{2}{c}{\textbf{P($Y=y|S=s$)}} \\
\midrule

 &$\downarrow y / s \rightarrow$ &Water &Land &4795 &Water &Land \\
\cmidrule(l){3-4}
\cmidrule(l){6-7}
Waterbirds &Water &3498 (73.0\%) &184 (3.8\%)  &3682 (76.8\%) &98.4\% &14.8\%\\
           &Land  &56 (1.2\%)   &1057 (22.0\%) &1113 (23.2\%) &1.6\%  &85.2\%\\
\midrule

 & &Female &Male &162770 &Female &Male \\
\cmidrule(l){3-4}
\cmidrule(l){6-7}
CelebA & Non-blonde &71629 (44.0\%) &66874 (41.1\%) &138503 (85.1\%) &75.8\% &98.0\%\\
       & Blonde     &22880 (14.1\%) &1387 (0.8\%)  &24267 (14.9\%)  &24.2\% &2.0\%\\

\bottomrule
\end{tabular}
\end{small}
\end{center}
\caption{Label and group details for worst-group accuracy benchmarks. 
These datasets have both label and group imbalances. The final columns calculate how the class probabilities shift when conditioning on the spurious attribute $s$.
The datasets exhibit a class imbalance which contributes to a large
group imbalance. For example, less than 15\% of the dataset has the label ``blonde''. A spurious correlation is created as 24\% of females are blonde while only 2\% of males are blonde. 
}

\label{tab:datasets}
\end{table*}

\subsection{MID Filtering}
\label{apn:filtering}

\begin{figure}[h]
\centering
    \includegraphics[width=\linewidth]{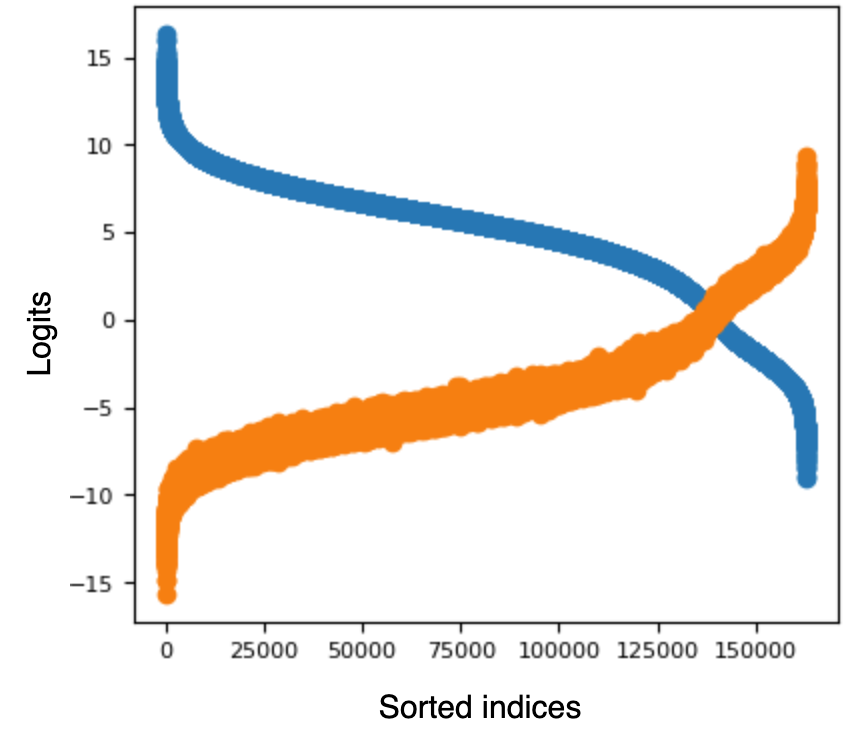} 
\caption{Logits in descending order for the class non-blonde (in blue), and blonde logits (in orange). } 
\label{fig:lofits_sorted}%
\end{figure}

\begin{figure*}[h]
\centering
    \includegraphics[width=\linewidth]{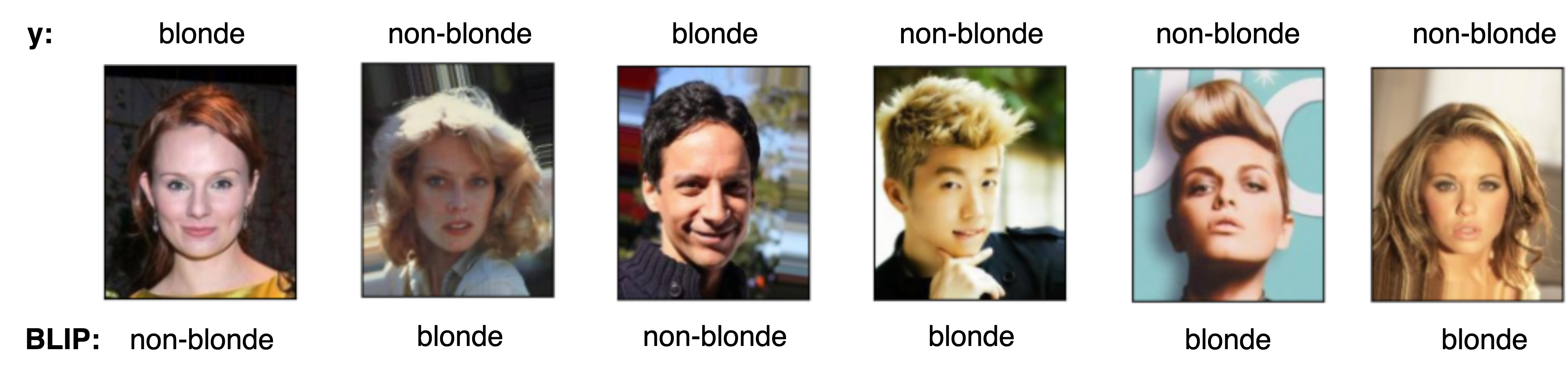} 
\caption{Middle logit images with y labels on top disputed by BLIP labels shown below each image. } 
\label{fig:mislabels}%
\end{figure*}

For CelebA, we took 3,839 points out of 162,770 (2,000 logit intercept points for each class, meaning 3,839 unique points as there were overlapping points). \cref{fig:lofits_sorted} shows the logits sorted in descending order for the non-blonde class. 
Of these, the ERM model misclassified 1,137 points (out of a total of 2063 misclassified points. That is, the selected data contains 55.1\% of all the ERM model's errors). 
Of the 3,839 logit intercept points, many of these points may appear ambiguous with respect to the label, or may have a wrong label. 
To handle mislabels and ambiguously labelled data, we apply BLIP in a VQA setting by asking ``Is this person blonde?'' for each image to obtain a ``yes'' or ``no'' answer. 
\cref{fig:mislabels} shows a sample of randomly selected images with disputed labels. The first four images on the left are clear mislabels. The two rightmost images contain celebrities with a debatable hair colour. 
In both scenarios, it is apparent why the model would be confused by such images. 
In total, BLIP disputed the label for 1679 points, leaving 2160 points remaining after step 2 in MID.

\subsection{MID Clusters and Retraining Details}
\label{apn:clustering}

\begin{figure*}[h]
\centering
    \includegraphics[width=\linewidth]{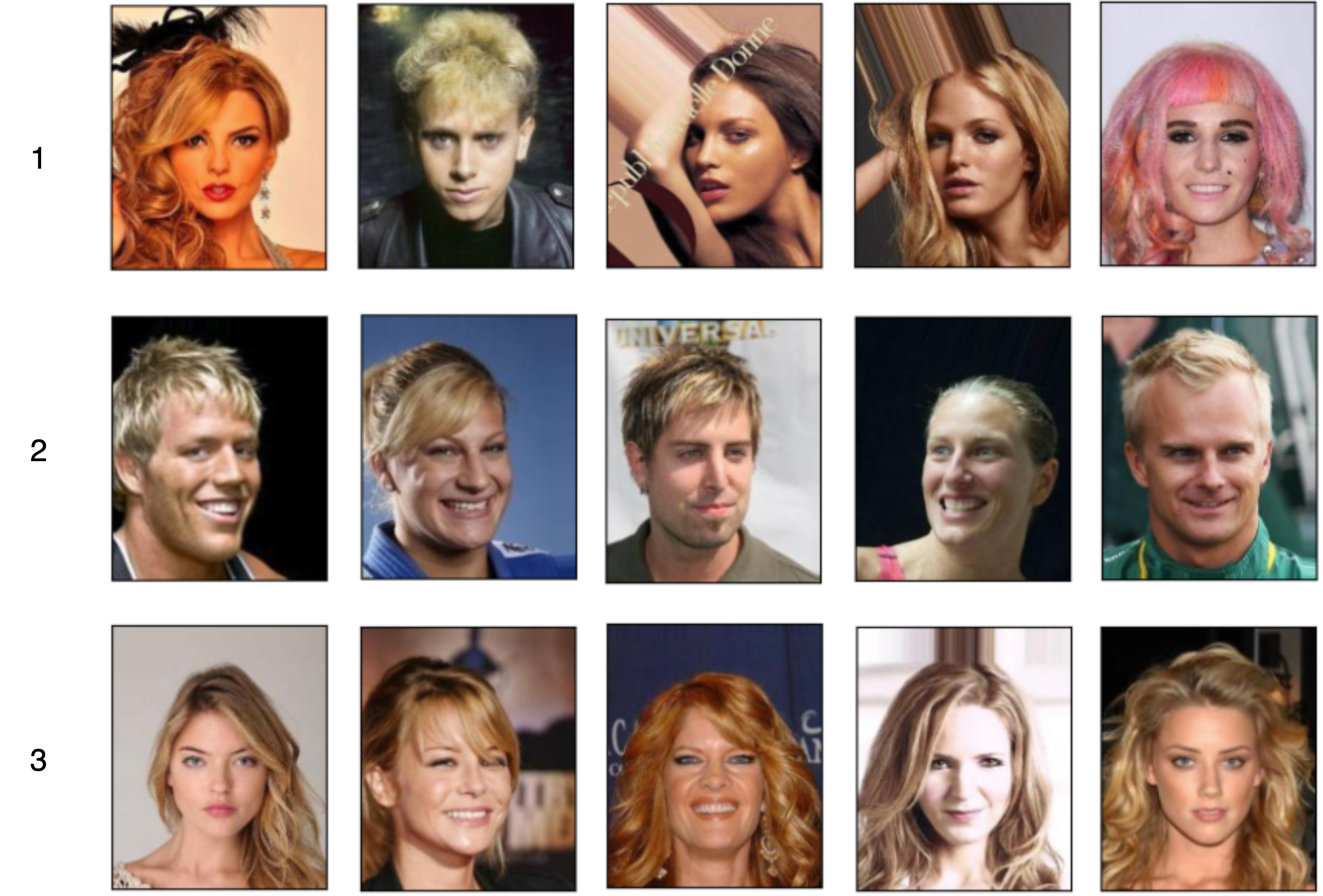} 
\caption{Group composition of k-means clusters for $k=3$ on the filtered CelebA dataset.} 
\label{fig:cluster_composition}%
\end{figure*}

\begin{figure}[h]
\centering
    \includegraphics[width=\linewidth]{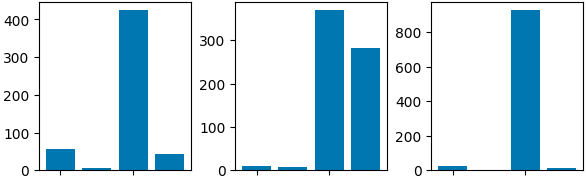} 
\caption{Group composition of k-means clustering on the filtered CelebA dataset.} 
\label{fig:cluster_w_filter}%
\end{figure}

\paragraph{K-means cluster analysis:}
K-means was applied using the scikit learn package~\cite{pedregosa2011scikit} starting with $k$ = 2 as described in \cref{sec:mid} with default settings besides a fixed random seed 0, 5 initial centroids, and a maximum number of iterations set to 1000. 
A manual inspection of 50 images from each cluster was allowed to establish if the clusters consisted of data that was reasonable for the model to exhibit uncertainty about. \cref{fig:cluster_composition} shows examples of images from each of the three clusters. Cluster 1 appears to consist of unusual images, such as strange lighting, unusual artefacts and hair colours that the model is likely not too familiar with. For these reasons, it is not unreasonable that the model might struggle with classifying these images. Cluster 2 contains many blonde males, but also some blonde females. Studying the pattern led us to summarise this cluster by containing masculine features such as 
a larger chin (male or female), short hair, or tightly tied back hair. Since the images inspected were all clearly blonde, the model's poor performance on this cluster is less understandable. 
Further, as discussed above, an apparent pattern is visible. Therefore, we marked this cluster for retraining. 
Cluster 3 is composed of images of female celebrities with hair colour that is between blonde and brunette, or auburn coloured hair. We believe this cluster reflects an understandable level of uncertainty expressed in the logits. 
\cref{fig:cluster_w_filter} shows the group compositions of the cluster, which our method does not use, but we include for demonstration purposes. 
The groups represent the combination of label and spurious attribute. 
Group 1 is the combination of non-blonde and female attributes, group 2 non-blonde males, group 3 blonde females, and group 4 is the minority group consisting of blonde males. 
As discussed above, cluster 2 contains mainly images of blonde celebrities. The corresponding middle chart in \cref{fig:cluster_w_filter} shows that this cluster contains many blonde males and blonde females. 
We note that we began clustering with 2 clusters and found good results for 3 clusters. However, we repeated analysis for 4 clusters to check if a pattern emerged. We found similar results, with the first cluster described above roughly splitting in two. 

\paragraph{Selecting data for retraining and manual inspection:}
The data used for retraining is simply taken from the clusters where poor model performance on cluster labels is deemed inappropriate (e.g., it is unacceptable to perform poorly in classifying the hair colour of people with masculine type features) and not clusters where poor performance is deemed reasonable (e.g., a cluster containing celebrities with in between hair colours). 
Regarding manual inspection and automating our method, the cluster labelling component can be done without human input using an appropriate VLM if available for the application. 
Automating the inspection of what is unusual about the clusters or if poor performance is acceptable 
is left to future work. 
This is because the method allows for the spurious correlation to not be preconceived (however, if the spurious correlation is preconceived, a VLM can again do this step), and reasoning about whether poor performance on a cluster is appropriate or not requires common sense, and domain knowledge in some cases. 

We suggest that for many applications an LLM may be able to assist with this component of the method (e.g., for the cluster of dark blonde and light brown-haired females, one could ask an LLM if it is reasonable for a model to have uncertainty classifying this type of hair as blonde or non-blonde?). 
However, a full study of the applicability of LLMs to automate this type of common sense human analysis should be undertaken before trusting a model with this task for many applications. 
Not requiring preconceived ideas about spurious attributes is a key feature of our work which distinguishes it from methods requiring group label information, which implicitly assume a preconceived spurious attribute. In many cases, we may not know which concepts are entangled with each other. 
For now, having a human in the loop to some extent may be wise to avoid failures for many applications where the spurious correlation is not preconceived as is allowed for in our MID experiments. Fortunately, given that MID significantly reduces the amount of data to be inspected, the human workload should be greatly reduced with inspection required for just a few images from each cluster. 

\paragraph{MID retraining:}
We train the regularisation parameter for the logistic regression reweighting on one-half of the data from poorly performing selected clusters (Step 4). We test for values in the range \{1.0, 0.7, 0.3, 0.1, 0.07, 0.03, 0.01\} and find a strength of 1.0 is the best-performing value for both CelebA and Waterbirds. We then select this optimal value that leads to the best accuracy and train on the available validation data with additional randomly selected training set data so that the model does not ``forget'' the majority groups leading to a poorer overall accuracy in favour of a higher worst-group accuracy. 

\subsection{MID Ablation Experiments}
\label{apn:ablations}

\begin{figure}[h]
\centering
    \includegraphics[width=\linewidth]{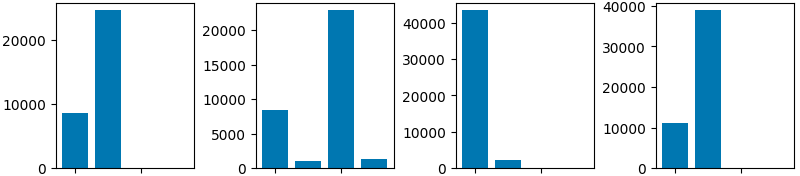} 
\caption{Group composition of k-means clustering on the entire CelebA dataset.} 
\label{fig:cluster_wo_filter}%
\end{figure}

In this section, we demonstrate the importance of taking only mid-level logits through the absence of Step 2 in our method, MID, described in \cref{sec:mid} for the CelebA dataset. 
We investigate the need to narrow our focus to the middle logit intercept data, we repeat k-means clustering (step 3 of our method) without the preceding filtering step (Step 2). 
\cref{fig:cluster_wo_filter} shows the resulting cluster compositions by label and spurious attribute. 
The pattern that emerges from analysing the clusters is roughly group 1 contains non-blonde celebrities with various backgrounds, group 2 consists of blonde and light-haired non-blondes, group 3 contains brunette celebrities with long hair, and group 4 contains non-blonde celebrities with short hairstyles. None of the clusters were found to reveal a spurious pattern in the data. This finding is in line with Sohoni \etal~\cite{sohoni2020no}. 
We further investigate the need to select specific clusters that the model should perform well on, rather than just a random cluster. We find that applying MID to random clusters does not necessarily improve model performance, and can negatively impact model performance in some cases. E.g., retraining the model on the cluster of photos of celebrities with hair that is between blonde and brown, does not help with performance and certainly not the WGA. This cluster reflects a reasonable model uncertainty as a human could easily spot that classifying these samples is difficult and that this cluster may also reflect some label inconsistency. 

\end{document}